\documentclass[lettersize,journal]{IEEEtran}
\usepackage{amsmath,amsfonts}
\usepackage{array}
\usepackage[caption=false,font=normalsize,labelfont=sf,textfont=sf]{subfig}
\usepackage{textcomp}
\usepackage{stfloats}
\usepackage{url}
\usepackage{verbatim}
\usepackage{graphicx}
\usepackage{cite}
\usepackage{multirow}
\usepackage{tabularx,booktabs}
\newcolumntype{C}{>{\centering\arraybackslash}X} 
\usepackage[dvipsnames]{xcolor}
\usepackage{longtable}
\usepackage{makecell, cellspace, caption}
\usepackage{array}
\usepackage{subcaption}
\usepackage{cuted}
\usepackage{multirow}
\usepackage[utf8]{inputenc}
\usepackage{tabularx}
\usepackage{array}
\usepackage{amssymb}
\usepackage{makecell}
\usepackage[margin=2cm]{geometry}
\usepackage{adjustbox}
\usepackage{multirow}
\usepackage{tabulary,booktabs}
\usepackage{ragged2e}
\usepackage{amsmath}
\usepackage{booktabs}
\usepackage{array}
\newcolumntype{L}{>{$}l<{$}}
\newcolumntype{C}{>{$}c<{$}}
\newcolumntype{R}{>{$}r<{$}}

\usepackage{amssymb}
\usepackage{pifont}
\usepackage{graphicx}
\usepackage[caption=false]{subfig}
\usepackage{hyperref}
\hypersetup{colorlinks,allcolors=black}
\usepackage{flushend}
\usepackage{mathtools}
\usepackage[linesnumbered,lined]{algorithm2e}
\makeatletter
\renewcommand{\@algocf@capt@plain}{above}
\RestyleAlgo{ruled}
\usepackage{setspace}
\SetKw{KwParam}{Params:}
\SetKw{KwEnsure}{Ensure:}
\SetKw{KwSet}{Set}
\SetKwComment{KwComment}{/* }{ */}
\makeatother
\begin{document}

\title{SSNet: Saliency Prior and State Space Model-based Network for Salient Object Detection in RGB-D Images}

\author{Gargi Panda, Soumitra Kundu, Saumik Bhattacharya, Aurobinda Routray, \IEEEmembership{Member, IEEE}
\thanks{Gargi Panda and Aurobinda Routray are with the Department of EE, IIT Kharagpur, India
(email: pandagargi@gmail.com; aroutray@ee.iitkgp.ac.in)}
\thanks{Soumitra Kundu is with the Rekhi Centre of Excellence for the Science of Happiness, IIT Kharagpur, India (e-mail: soumitra2012kbc@gmail.com).}
\thanks{Saumik Bhattacharya is with the Department of E\&ECE, IIT Kharagpur, India
(email: saumik@ece.iitkgp.ac.in)}
}

\markboth{Journal \LaTeX\  }%
{Shell \MakeLowercase{\textit{et al.}}: A Sample Article Using IEEEtran.cls for IEEE Journals}


\maketitle

\begin{abstract}
Salient object detection (SOD) in RGB-D images is an essential task in computer vision, enabling applications in scene understanding, robotics, and augmented reality. However, existing methods struggle to capture global dependency across modalities, lack comprehensive saliency priors from both RGB and depth data,  and are ineffective in handling low-quality depth maps. To address these challenges, we propose SSNet, a saliency-prior and state space model (SSM)-based network for the RGB-D SOD task. Unlike existing convolution- or transformer-based approaches, SSNet introduces an SSM-based multi-modal multi-scale decoder module to efficiently capture both intra- and inter-modal global dependency with linear complexity. Specifically, we propose a cross-modal selective scan SSM (CM-S6) mechanism, which effectively captures global dependency between different modalities. Furthermore, we introduce a saliency enhancement module (SEM) that integrates three saliency priors with deep features to refine feature representation and improve the localization of salient objects. To further address the issue of low-quality depth maps, we propose an adaptive contrast enhancement technique that dynamically refines depth maps, making them more suitable for the RGB-D SOD task. Extensive quantitative and qualitative experiments on seven benchmark datasets demonstrate that SSNet outperforms state-of-the-art methods.
\end{abstract}

\section{Introduction}
\IEEEPARstart{S}{alient} object detection (SOD) is a fundamental computer vision task that aims to identify the most prominent object in a natural scene, drawing maximum human attention. It is commonly used to facilitate subsequent vision tasks such as content-aware image editing \cite{editing}, image cropping \cite{cropping}, segmentation \cite{segmentation}, tracking \cite{tracking}, etc. RGB SOD methods primarily depend on contrast variations to distinguish salient objects from their backgrounds. But when the foreground and background share similar appearances, SOD performance significantly degrades. To address this limitation, researchers have turned to RGB-D SOD, which integrates depth (D) data with RGB images, imitating human-like stereoscopic vision for enhanced object differentiation \cite{des}. This fusion of color and depth information provides a more comprehensive overview of the scene, improving SOD accuracy. In the era of deep learning (DL), RGB-D SOD methods generally use the convolutional neural network (CNN) \cite{cdnet, bianet, hainet, midd, dcf, moadnet, siamese, cirnet, hidanet, fastersal, rd3dplus} or vision transformer (ViT)-based methods \cite{caver, swinnet, hinet, lafb, vstplus} to encode and fuse multi-modal features for saliency map prediction. ViT-based methods, in particular, utilize self-attention and cross-attention mechanisms to model long-range dependencies, achieving superior performance over CNN-based techniques. However, the attention mechanism in the transformer has quadratic complexity with the input size, leading to significant computational overhead, especially in SOD tasks that require high-resolution feature maps \cite{vmamba}. To tackle this issue, ViT-based approaches \cite{caver, swinnet, hinet, lafb, vstplus} divide feature maps into non-overlapping windows and apply the attention mechanism to each window independently. While this strategy reduces computational costs, it limits the model’s ability to capture global dependencies across the entire feature map. However, modeling the global dependency is crucial for precise salient object detection, as it enhances the network’s ability to understand spatial relationships and semantic connections between different regions of an image \cite{multiGT}. Inspired by advances in natural language processing (NLP), researchers have explored alternative architectures to improve efficiency without sacrificing performance. Recently, a state space model (SSM)-based approach, has demonstrated remarkable success in the NLP domain in modeling long-range dependencies with linear complexity \cite{mamba}. Encouraged by its effectiveness, some works have adapted its selective scan SSM (S6) mechanism for vision tasks \cite{visionmamba, vmamba, vmunet2}. One very recent RGB-D SOD method \cite{mambasod2} has applied the S6 mechanism for capturing global dependency across high-resolution feature maps. However, despite these advancements, S6 lacks a mechanism to model long-range dependencies across different modalities, limiting its ability to fully exploit the complementary nature of RGB and depth information. Moreover, \cite{mambasod2} uses a computationally expensive cross-scanning mechanism \cite{vmamba} to apply S6 in four directions of the image. These challenges motivate us to design novel SSM-based modules to efficiently capture global dependency across intra and inter-modality images.

Beyond architectural improvements, another critical aspect of RGB-D SOD is the quality of the depth map itself. Depth images play a crucial role in differentiating foreground from background, but in real-world scenarios and commonly used datasets, they often suffer from noise, missing details, or low resolution due to limitations in data acquisition \cite{cdnet}. Some existing methods address this issue by either discarding low-quality depth maps \cite{impDepth1,impDepth2,dpanet} or estimating high-quality depth from RGB images \cite{cdnet, delving}. 
Additionally, some RGB-D SOD methods leverage depth histograms as prior information to enhance the discriminative power of deep features \cite{dsa2f,hidanet}. While this helps improve object segmentation, these methods rely solely on depth contrast. 
This gap motivates us to integrate more comprehensive saliency-based priors that consider both RGB and depth cues, further improving the network’s ability to discriminate salient objects from complex or ambiguous backgrounds. 

Motivated by these challenges, we introduce SSNet, a novel saliency-prior and SSM-based network for the RGB-D SOD task. SSNet integrates saliency domain knowledge-based priors with deep features and employs an SSM-based multi-modal multi-scale decoder module (M\textsuperscript{2}DM) to predict the saliency map. Specifically, we propose a novel cross-modal selective scan SSM (CM- S6) mechanism to capture the global dependency between different modalities. By leveraging the S6 and CM-S6 mechanisms, we design a self-modality global feature block (SGFB) and a cross-modality global feature block (CGFB) to mine the global context in intra and inter-modalities. Besides, to address the direction sensitivity of S6 and CM-S6 mechanisms, we introduce a sequence flipping technique.  In addition, we propose a saliency enhancement module (SEM) to integrate saliency priors with deep features to improve foreground-background separation. Moreover, to address the depth quality issues, we propose an adaptive contrast enhancement (ACE) scheme to make the depth maps more suitable for the SOD task. By integrating M\textsuperscript{2}DM, SEM, and ACE, SSNet effectively tackles global dependency modeling across modalities, depth quality enhancement, and saliency-aware feature learning, leading to superior RGB-D SOD performance. Our key contributions can be summarized as:

\begin{enumerate}
\item We propose a novel SSM-based CM-S6 mechanism to capture global dependency across multi-modalities.
\item We introduce an ACE scheme to improve the quality of depth maps for better SOD performance.
\item We develop a novel SEM that integrates saliency-based priors with deep features, enhancing their discriminative capability in detecting salient regions. 
\item We design SGFB and CGFB, based on S6 and CM-S6 mechanisms respectively, to capture the global context in intra and inter-modalities. Leveraging SGFB, CGFB, ACE, and SEM together, we develop our proposed SSNet for the RGB-D SOD task.
\item Extensive experiments across seven benchmark datasets demonstrate that SSNet outperforms SOTA methods in the RGB-D SOD task.
\end{enumerate}

The rest of this paper is structured as follows: we review related works in RGB-D SOD and SSM in computer vision in Section \ref{sec2}. Section \ref{sec3} details our framework and core components of SSNet. Section \ref{sec4} presents experimental evaluations, followed by the conclusion in Section \ref{sec5}.
\section{Background}
\label{sec2}
\begin{figure*} 
    \centering
  \includegraphics[width=0.75\linewidth]{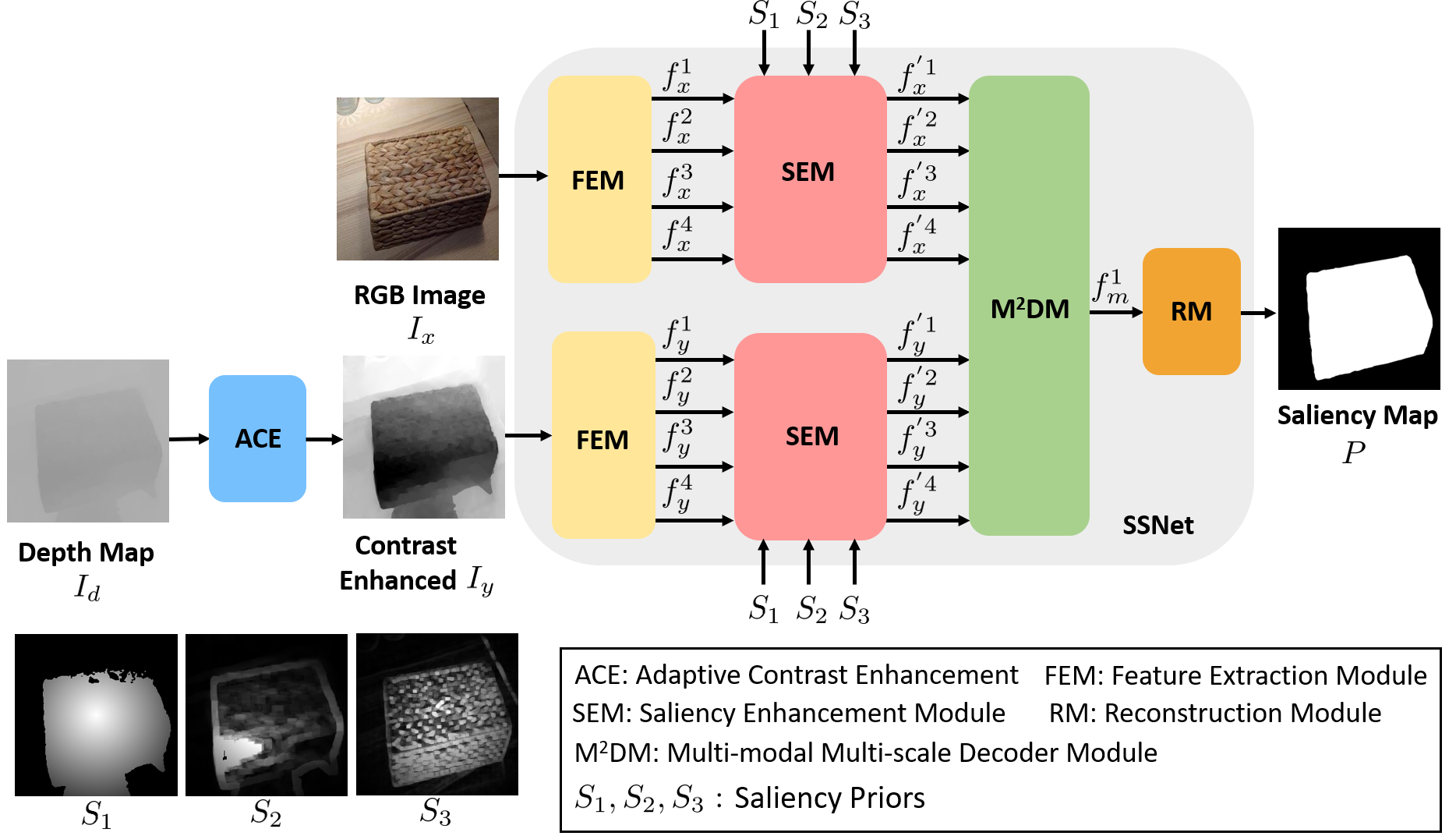}
  
  \caption{Overall framework and architecture of SSNet.}
  
  \label{fig2} 
\end{figure*}
\subsection{SOD in RGB-D Images}
Recent RGB-D SOD methods \cite{cdnet, bianet, hainet, midd, dcf, sip, swinnet, moadnet, siamese, cirnet, hidanet, hinet, caver, lsnet, disentangled, fastersal, rd3dplus, lafb, vstplus, emtrans} primarily focus on DL techniques. The mainstream approach is to encode multi-modal multi-scale features from the RGB and depth images and design different fusion schemes for predicting the final saliency map. CDNet \cite{cdnet} designed a two-stage fusion scheme to combine low-level and high-level depth features with the features from RGB images. BiANet \cite{bianet} proposed a bilateral attention module to capture the cross-modal features in both the foreground and background regions. RD3D+ \cite{rd3dplus} proposed a 3D CNN-based cross-modal fusion module for the RGB-D SOD task. These methods predominantly rely on the convolution operation, which excels at capturing local context but struggles with modeling global dependency. To overcome this limitation, many methods \cite{swinnet, hinet, caver, disentangled, lafb, vstplus, emtrans} have incorporated transformer self-attention and cross-attention mechanisms in designing networks for the RGB-D SOD task. CAVER \cite{caver} introduced a mixed-view attention module, that applies the attention mechanism in both the spatial and channel dimensions. Liu et al. \cite{vstplus} proposed task-related tokens to effectively mine complementary information in RGB-D images. LAFB \cite{lafb} designed an adaptive decoder module that dynamically selects the fusion scheme for the multimodal features. EM-Trans \cite{emtrans} introduced an edge-aware transformer network that considers edge information of the salient region to refine salient object boundaries. Though transformer-based methods show superior performance compared to CNN-based methods, the transformer attention mechanism has quadratic complexity with the input size. This makes a substantial computational overload when applying the attention mechanism to high-resolution feature maps in the SOD task. To address this computational issue, the transformer-based methods divide the feature maps into non-overlapping windows and apply the attention mechanism to each window independently. However, this window-based method hinders the network to model the global dependency, which is important in the RGB-D SOD task \cite{multiGT}. One very recent method \cite{mambasod2} has applied the SSM-based mamba mechanism \cite{mamba} to capture global context. In \cite{mambasod2}, the features from RGB-D images are correlated and the mamba mechanism is applied to capture cross-modal dependency. However, relying solely on simple correlation is insufficient for effectively mining dependency across different modalities \cite{caver, disentangled, vstplus}. In the next subsection, we discuss the SSM method in computer vision.

In addition to effective feature fusion, the quality of the depth map is crucial for accurately identifying salient objects. Depth maps become less effective when the contrast between salient objects and the background is low. However, in real-world scenarios and commonly used datasets, depth maps often suffer from quality degradation due to limitations in data acquisition, reducing their reliability \cite{cdnet}. Some previous methods have addressed this issue by either discarding low-quality depth maps \cite{impDepth1,impDepth2,dpanet, sip} or estimating depth maps directly from RGB images \cite{cdnet, delving}. Instead of discarding or synthesizing depth information, we propose an adaptive contrast enhancement scheme that refines the original depth map, making it more effective for the RGB-D SOD task.

\subsection{SSM in Computer Vision}
State Space Models (SSMs), with their linear complexity, have emerged as a strong alternative to transformers for modeling long-range dependency. The recently introduced SSM-based architecture, mamba \cite{mamba}, has demonstrated superior performance over transformers in capturing long-range context in the NLP domain. Mamba’s key innovation lies in its selective scan SSM (S6) mechanism, where parameters are input-dependent (selective), and the output is computed recursively (scan). Mamba enables an effective transformation of long sequences by utilizing a hardware-aware algorithm to efficiently solve recurrence equations. Following its success in NLP, mamba has been increasingly adapted to computer vision tasks like image classification, image segmentation \cite{visionmamba, vmamba, vmunet2}, image restoration \cite{mambair, VmambaIR}, action recognition \cite{graphsimba}, object tracking \cite{TrackingMamba}. A very recent RGB-D SOD study \cite{mambasod2} also applied the S6 mechanism to capture global dependency. However, S6 lacks a dedicated mechanism to model global dependency across different modalities, which is crucial for the multi-modal RGB-D SOD task. Additionally, a key challenge in applying the S6 mechanism to vision tasks is its inherent design for 1D sequential data, where sequence direction matters. When used in vision applications, images are flattened into 1D sequences before applying S6, but this approach fails to fully model long-range dependencies across different spatial directions. To address this limitation, VMamba \cite{vmamba} introduced a cross-scanning approach that applies the S6 mechanism in four directions. While effective, this method is computationally expensive \cite{es2d}. In our work, we propose a novel cross-modal selective scan SSM (CM-S6) mechanism that effectively captures global dependency between RGB and depth modalities. Furthermore, to address the direction sensitivity of the recursive scan in the S6 mechanism, we introduce a simple sequence-flipping technique that enables efficient modeling of global dependency in two directions while maintaining computational efficiency.

\subsection{Including Prior Knowledge in RGB-D SOD}
In RGB-D SOD, some approaches incorporate prior knowledge from depth maps into deep neural networks to enhance saliency detection. Zhao et al. \cite{contrastprior} introduced a contrast prior loss to improve the distinction between foreground and background in an enhanced depth map for the SOD task. Other methods \cite{dsa2f,hidanet} decompose depth maps into multiple regions based on histograms and integrate these with deep RGB-D features. However, these existing methods rely solely on depth contrast to compute priors. In contrast, our approach considers both RGB-D contrast and two additional visual saliency-based assumptions to derive three saliency priors, which more accurately approximate the salient regions.

Based on the above-mentioned insights, we propose a novel approach that enhances global dependency modeling, refines depth map quality, and incorporates more effective saliency priors. In the next section, we introduce our proposed method in detail.

\section{Proposed Method}
\label{sec3}
\begin{figure}[!hbt] 
    \centering
  \includegraphics[width=0.9\linewidth]{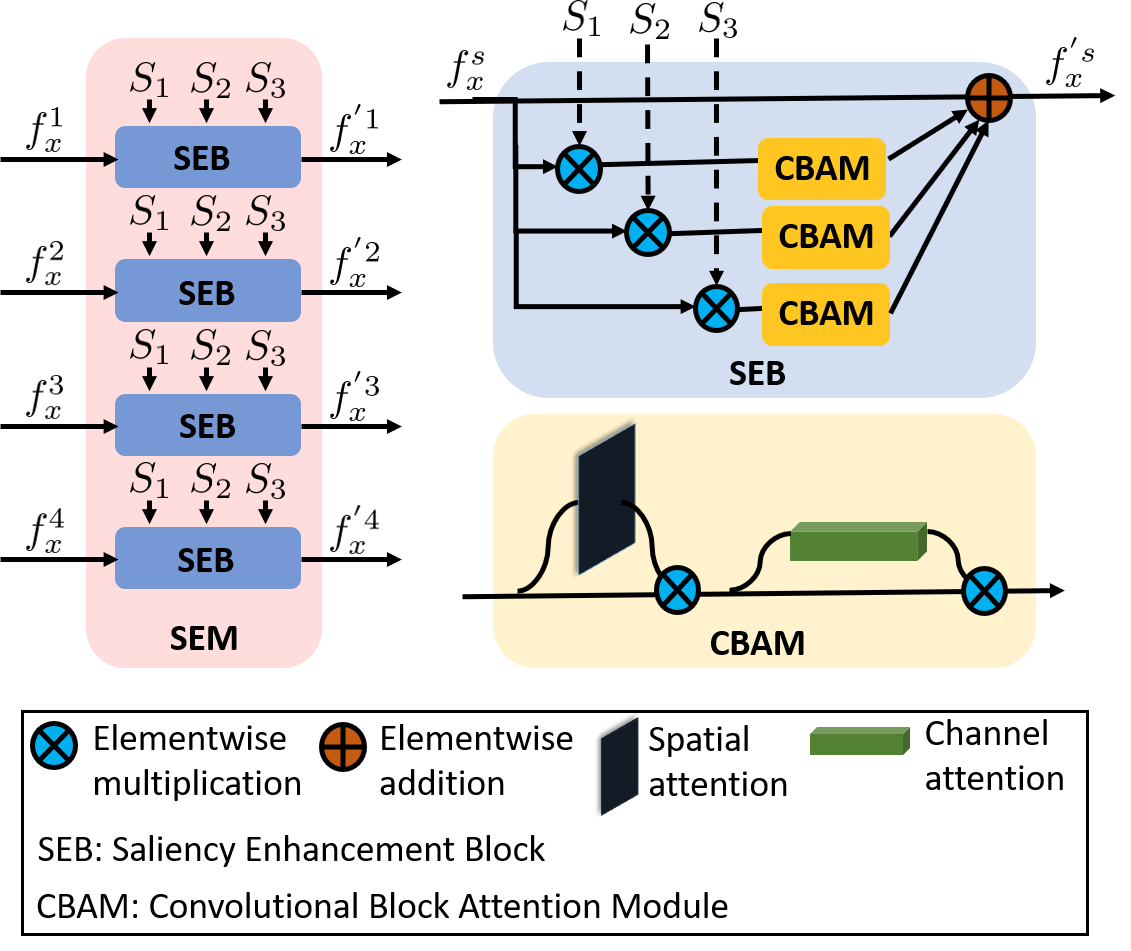}
  
  \caption{Structure of Saliency Enhancement Module (SEM).}
  
  \label{fig3} 
\end{figure}

\subsection{Overall  Framework}
\begin{figure*} 
    \centering
  \includegraphics[width=0.8\linewidth]{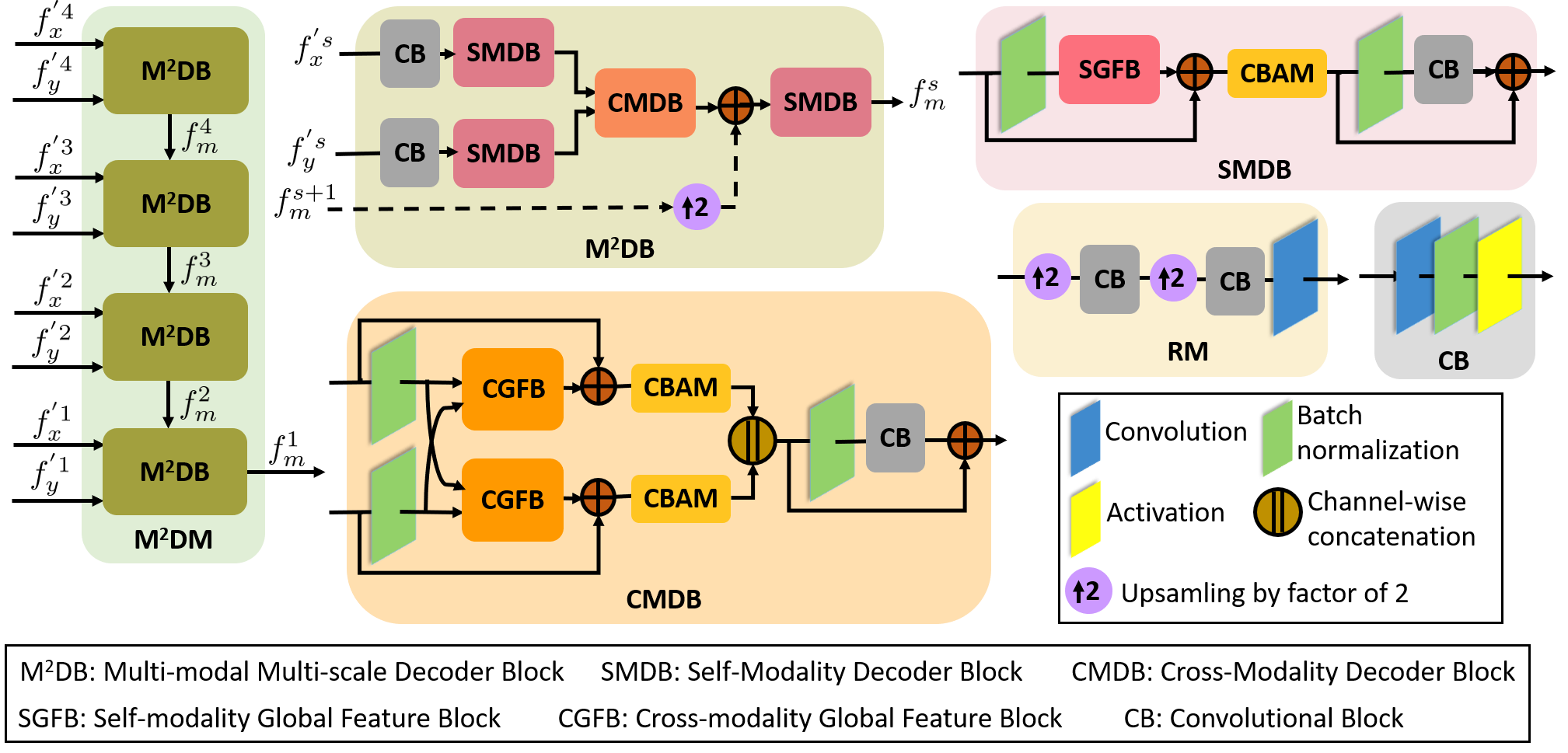}
  
  \caption{Structure of Multi-modal Multi-scale Decoder Module (M\textsuperscript{2}DM) and Reconstruction Module (RM).}
  \label{fig4} 
\end{figure*}

For the RGB-D SOD task, we consider an RGB image $I_{x}\in\mathbb{R}^{H, W,3}$, and its corresponding depth map  $I_{d}\in\mathbb{R}^{H, W,1}$, where $H$ is the image height and $W$ is the image width. Figure \ref{fig2} provides a visual representation of the overall framework. First, we process the depth map using an adaptive contrast enhancement (ACE) stage to obtain a contrast-enhanced depth map $I_{y}\in\mathbb{R}^{H, W,1}$. Then we input $I_x, I_y$ to SSNet, where two CNN backbones are utilized as feature extraction modules (FEM).  From the two FEMs, we get multi-scale features $f_x^s\in\mathbb{R}^{\frac{H}{4s},\frac{W}{4s}, sC}$, $f_y^s\in\mathbb{R}^{\frac{H}{4s},\frac{W}{4s}, sC}$ for the two modality images, where $s\in \{1,2,3,4\}$ and $C$ is feature channel dimension. Features $f_x^s$ from the RGB image contain color, texture, and appearance information, whereas features $f_y^s$ from the depth map are abundant with spatial structural information. These features $f_x^s,f_y^s$ are input to two saliency enhancement modules (SEM) to get $f_x^{'s},f_y^{'s}$, which have better discriminative capability in detecting the salient regions. Following the SEMs, we feed the features to a novel SSM-based multi-modal multi-scale decoder module (M\textsuperscript{2}DM) to get $f_m^1\in\mathbb{R}^{\frac{H}{4},\frac{W}{4}, D}$. This feature is fed to the reconstruction module (RM), whose structure is shown in Figure \ref{fig4}, to get the final saliency map $P\in\mathbb{R}^{H, W,1}$. 

\begin{table} [h!]
\centering
\includegraphics[width=\linewidth]{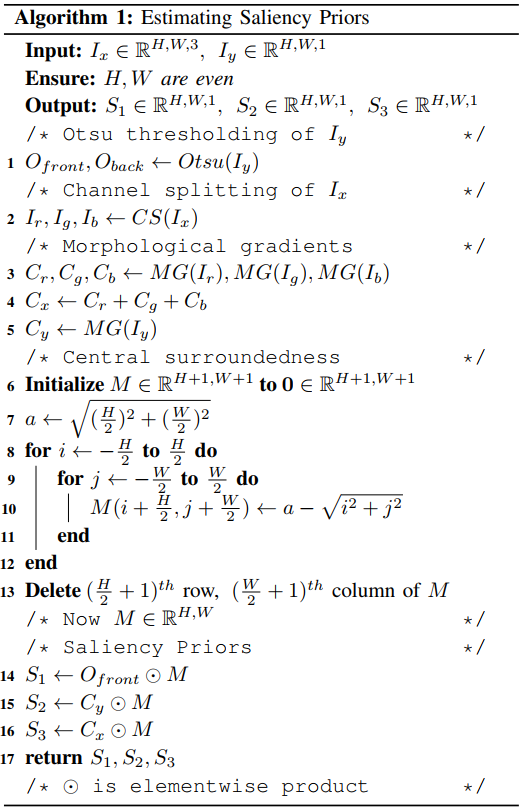}
\end{table}

\subsection{Adaptive Contrast Enhancement (ACE)}
The purpose of employing the ACE is to enhance the contrast of a low-quality depth map to make it more suitable for the RGB-D SOD task. Given the depth map $I_d$, we first calculate the bottom 1\% and the top 1\% of all pixel values as $I_{low}$ and $I_{high}$, respectively. The pixel values between $I_{low}$ and $I_{high}$ are linearly mapped between the range $[0, 1]$. The pixels with values less than $I_{low}$ are saturated to value $0$ and higher than $I_{high}$ are saturated to value $1$. Using this process, we get the enhanced image $I_y$. Figure \ref{fig2} shows an example of a low contrast depth map $I_d$ and the contrast-enhanced image $I_y$. We input this contrast-enhanced depth map $I_y$ in calculating saliency priors, and the feature extraction modules.

\begin{figure*} 
    \centering
  \includegraphics[width=0.75\linewidth]{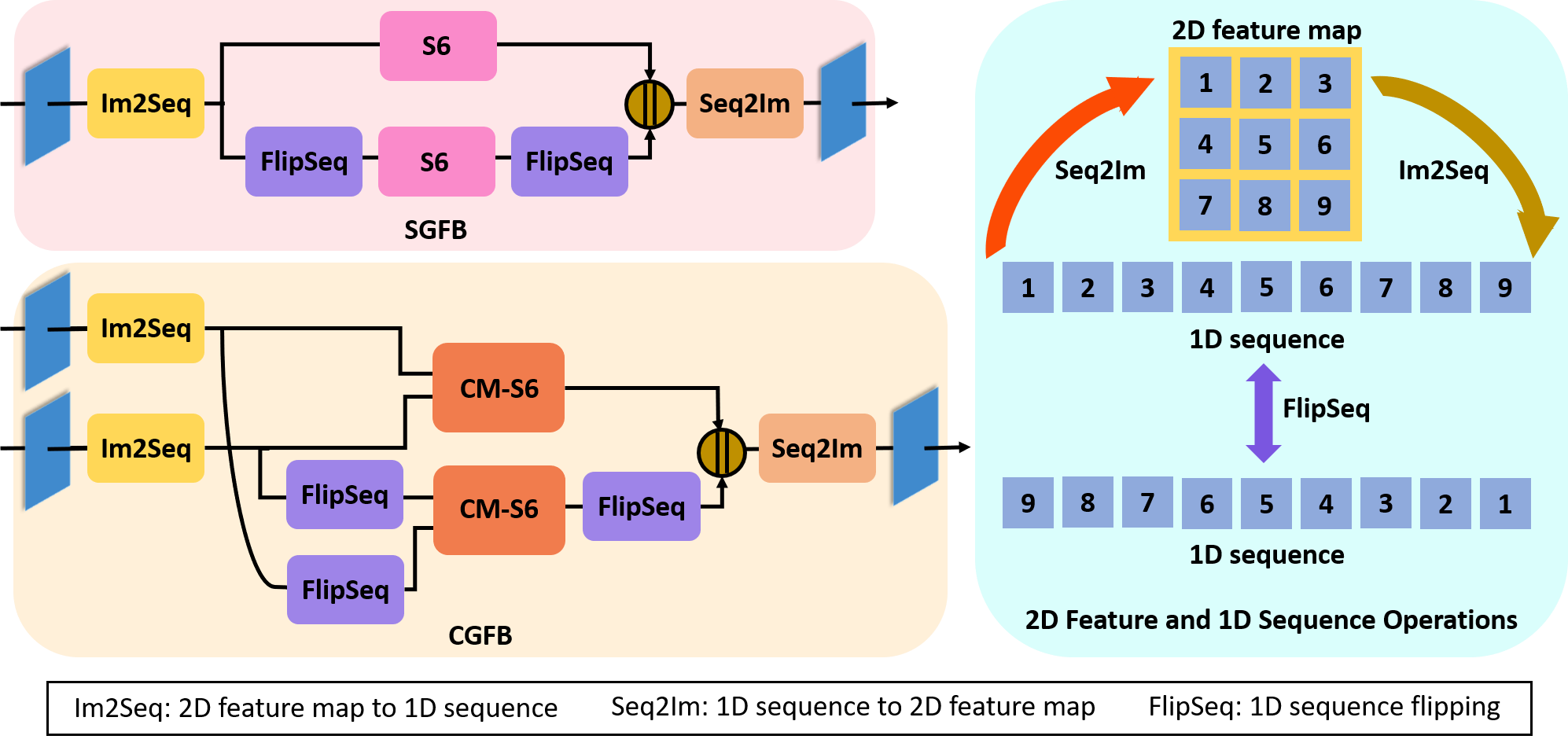}
  
  \caption{Architecture of Self-modality Global Feature Block (SGFB) and Cross-modality Global Feature Block (CGFB).}
  \label{fig5} 
\end{figure*}
\subsection{Saliency Enhancement Module (SEM)}
\label{sem}
SEM is mainly proposed to integrate prior knowledge related to visual saliency with the extracted deep features from RGB-D images, to enhance the features' discriminative capability in detecting salient objects. First, we estimate three easy-to-generate simple saliency priors $S_1, S_2, S_3$ based on the following assumptions: 
\begin{itemize}
\item The salient objects are most likely to be in front of their surroundings \cite{assump3}. 
\item The saliency of a region depends on the contrast in local neighborhoods \cite{assump1}.
\item In man-made photographs, salient objects are most often located near the center region, and away from the borders of the images \cite{assump2}.
\end{itemize}

Algorithm 1 illustrates the steps in estimating the saliency priors. First, we use the Otsu method \cite{otsu}, denoted as $Otsu (\cdot)$, to discretize the depth map $I_y$ in two different regions. Two masks $O_{front}$ and $O_{back}$ are generated. Higher pixel values in $O_{front}$ indicate the front regions in the depth image. To measure the contrast map of an image, we use the morphological gradient method \cite{morphological}, which we denote as $MG(\cdot)$. For the RGB image $I_x$, we first separate the three color channels using the channel splitting operation, denoted as $CS(\cdot)$. Then we estimate three separate contrast maps $C_r, C_g, C_b$ for the R, G, and B channels and add them to get a single contrast map $C_x$. For the depth image, the calculated contrast map is $C_y$. In the supplementary material, we explain the Otsu and the morphological gradient method in detail. To measure the central surroundedness, we estimate the mask $M\in\mathbb{R}^{H, W,1}$. Higher pixel values in $M$ indicate that the regions are close to the image center point. Finally, we calculate the saliency priors based on our estimated $O_{front}, C_y, C_x$, and $M$. Figure \ref{fig2} shows the estimated saliency priors $S_1, S_2, S_3$ for the input RGB-D image pair. As evident, the three saliency priors together can approximately identify the region as in the saliency map $P$. 

After estimating the saliency priors in the data preprocessing step, we integrate these priors with the multi-scale deep features using our proposed SEM. Figure \ref{fig3} shows the structure of SEM, which has four separate saliency enhancement blocks (SEB) for the four different scales. In each SEB, the feature $f_x^s$ is multiplied with the three saliency priors to highlight the salient regions. Then spatial and channel attention are applied in the convolutional block attention module (CBAM) \cite{cbam} to further refine the feature map. Finally, these enhanced features are aggregated using a residual connection.
\begin{table*} [h!]
\centering
\includegraphics[width=\linewidth]{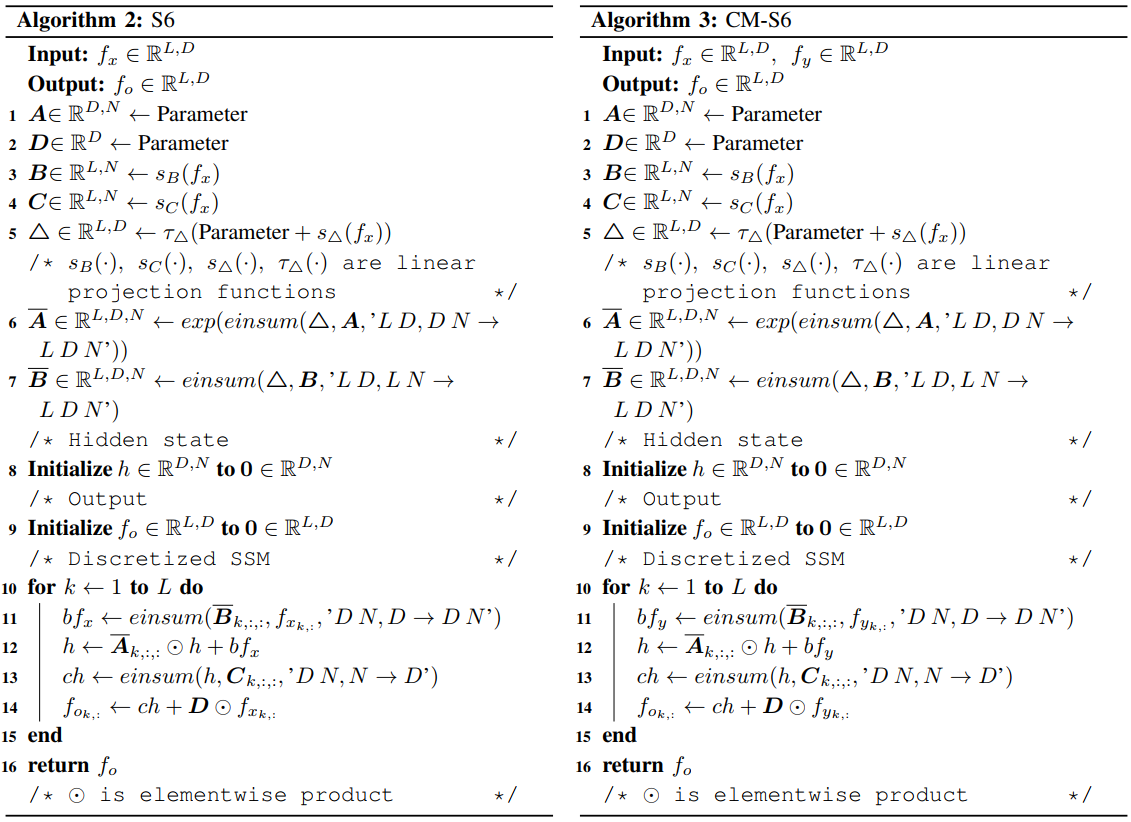}
\label{tab:sota}
\end{table*}
\subsection{Multi-modal Multi-scale Decoder Module (M\textsuperscript{2}DM)}

M\textsuperscript{2}DM is designed to integrate the multi-scale features from the two modality images. Figure \ref{fig4} depicts the architecture of M\textsuperscript{2}DM, which consists of four multi-modal multi-scale decoder blocks (M\textsuperscript{2}DB) for processing features at four different scales. For scale $s=4$, the M\textsuperscript{2}DB only processes the multi-modal features $f_x^{'4}$ and $f_y^{'4}$ to generate the feature $f_m^4$. For other three M\textsuperscript{2}DBs, along with $f_x^{'s}$ and $f_y^{'s}$, the feature $f_m^{s+1}$ from the higher scale is also processed. First the features $f_x^{'s}$ and $f_y^{'s}$ are fed to two convolutional blocks (CB) to capture the local context in the features. As shown in Figure \ref{fig4}, CB consists of a series of convolution, batch normalization, and activation layers. Then these local correlation-enhanced features are input to SSM-based self-modality decoder blocks (SMDB) to capture the intra-modality global context. Following this, one novel SSM-based cross-modality decoder block (CMDB) is applied to mine the inter-modal global dependency. Then, this cross-modal feature is added with the higher scale feature $f_m^{s+1}$ for scales $s=1,2,3$. Finally, one SMDB is applied to capture the long-range dependency in the feature. The details of SMDB and CMDB are discussed below.

\begin{table*} [h!]
\centering
\caption{Performance comparison to SOTA methods. FLOPs and runtime values are calculated under the setting of SOD from RGB-D images of resolution $384\times 384$. We highlight the best performances in \textcolor{red}{\textbf{red}} and second-best performances in \textcolor{blue}{\textbf{blue}} colors. $\uparrow$ and $\downarrow$ mean high and low values are desired, respectively.}
  \includegraphics[width=\linewidth]{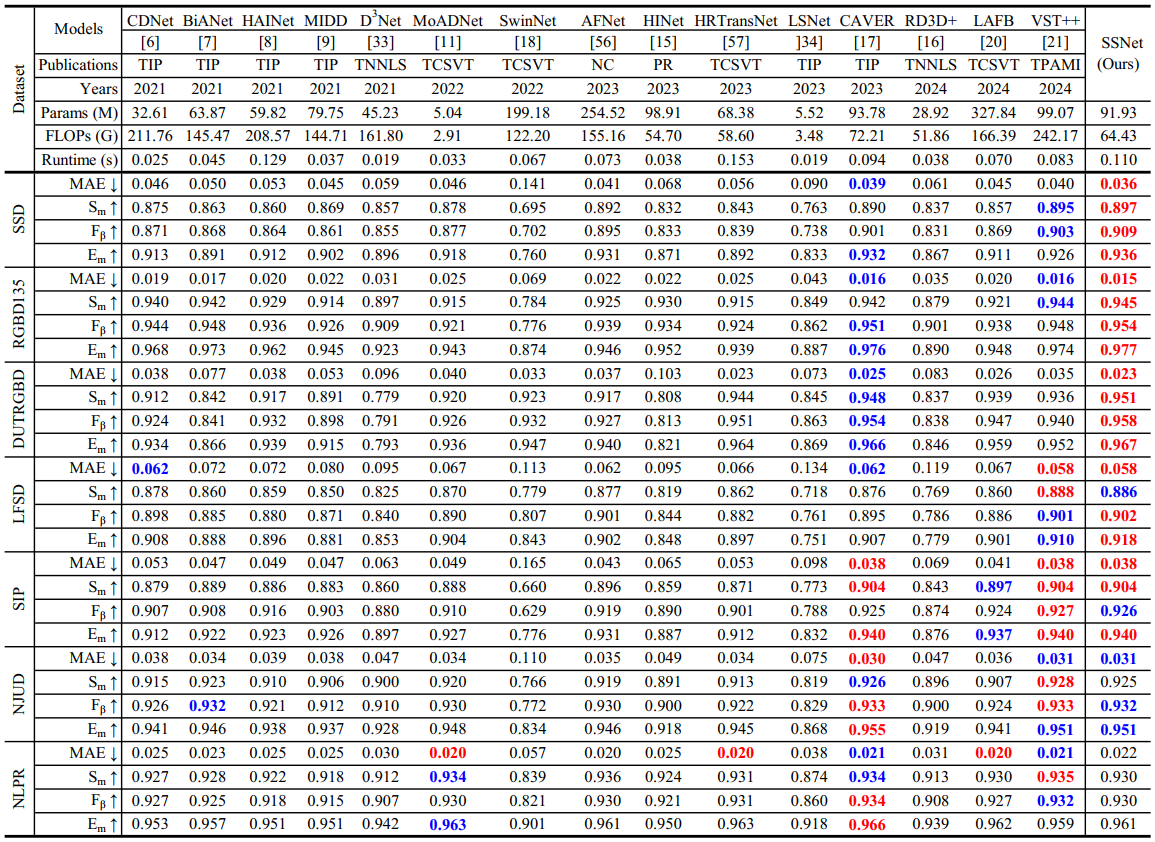}
\label{tab:sota}
\end{table*}
\begin{figure*} [h!]
\centering  \includegraphics[width=\linewidth]{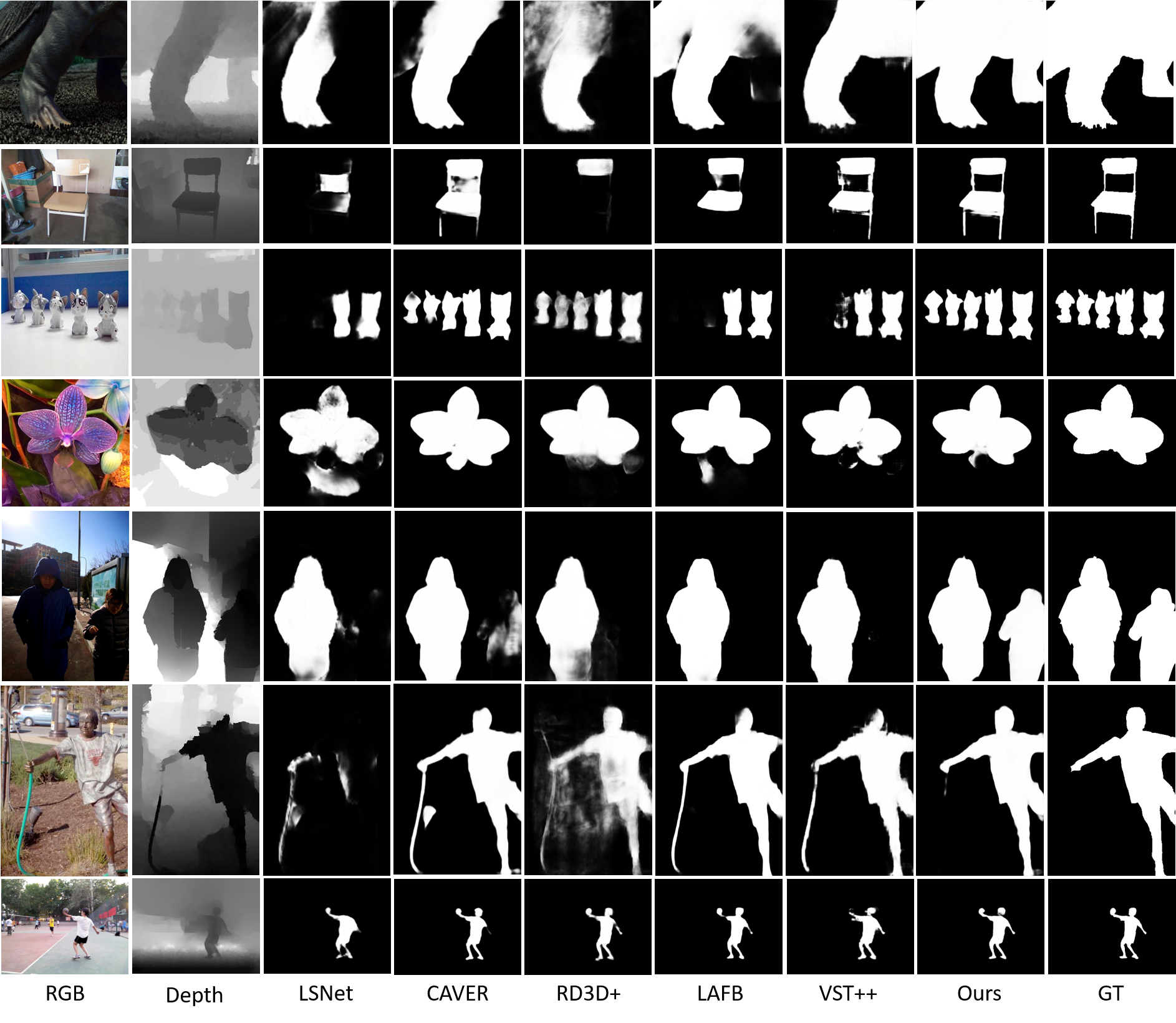}
\caption{Visual comparisons of SSNet to SOTA methods.}
\label{fig:sota_visual}
\end{figure*}

\subsubsection{Self-Modality Decoder Block (SMDB)}

In SMDB, first batch normalization (BN) is applied to the input feature, and then the self-modality global feature block (SGFB) is used to capture global dependency. This processed feature is used as a residual connection, which is further refined using CBAM. Following this step, BN and CB are applied sequentially as a residual connection to perform local correlation-based feature enhancement. 

Figure \ref{fig5} shows the structure of SGFB, which is mainly designed based on the S6 mechanism to capture global dependency in the same modality feature. First, one convolution layer is applied for linear projection, and then the 2D feature map is rearranged to a 1D sequence using the Im2Seq operation. Then, the S6 mechanism is applied. Algorithm 2 explains the S6 mechanism, which is based on a discretized SSM. In S6, the input sequence $f_x\in \mathbb{R}^{L,D}$ is transformed to output sequence $f_o \in \mathbb{R}^{L,D}$ through hidden state $h \in \mathbb{R}^{L,D}$ and the discretized state matrix 
$\boldsymbol{\overline{A}}\in \mathbb{R}^{L,D,N}$, input matrix 
$\boldsymbol{\overline{B}}\in \mathbb{R}^{L,N}$, output matrix 
$\boldsymbol{C} \in \mathbb{R}^{L,N}$, and feedthrough matrix $\boldsymbol{D} \in \mathbb{R}^{N}$. Here, $L$ and $D$ denote the input sequence length and channel dimension, respectively. The discretized parameters $\boldsymbol{\overline{A}}$ and $\boldsymbol{\overline{B}}$  are obtained by discretizing continuous time state matrix  $\boldsymbol{A}\in \mathbb{R}^{D, N}$ and input matrix $\boldsymbol{B}\in \mathbb{R}^{L, N}$ through timestep $\boldsymbol{\triangle} \in \mathbb{R}^{L, D}$. The notation $einsum(\cdot)$ is the Einstein summation operation. In S6, the time step and all the matrices are learnable parameters. For calculating $f_o$, the S6 mechanism uses the selective scan (SS) method \cite{mamba}, where the parameters 
$\boldsymbol{B}, \boldsymbol{C}, \boldsymbol{\triangle}$ are derived from the input feature $f_x$. The timestep parameter $\boldsymbol{\triangle}$ decides how much time to focus on the current input $f_{x_{k,:}}$. Deriving $\boldsymbol{\triangle}$ from $f_x$ makes the time spent on current input $f_{x_{k,:}}$ dependent on content (input $f_x$).
Moreover, by deriving $\boldsymbol{B}$ and $\boldsymbol{C}$  from $f_x$, we control how much content can be in the context (hidden state $h$) and how much context can be in the output $f_o$ based on the content. Thus, the model dynamics become dependent on model input $f_x$. Finally, the output $f_o$ is calculated efficiently by solving a recurrence equation using a hardware-aware algorithm. It is worth noting that, unlike the transformer-based methods dividing the feature map into non-overlapping windows, we can rearrange the entire 2D feature map as a 1D sequence and apply the S6 mechanism.  

However, due to the recurrence in SSM shown in Algorithm 2, the output $f_o$ becomes sensitive to the direction of the input sequence $f_x$. We propose to use a simple sequence flipping technique to apply the S6 mechanism in two directions of the 1D sequence. As shown in Figure \ref{fig5}, we also flip the sequence, apply the S6 mechanism, and then again flip the output sequence, which is concatenated channel-wise. This way, we model the global dependency in two directions of the feature map. Finally, the concatenated sequence is rearranged back to a 2D feature map and one convolution layer is applied for linear projection.

\subsubsection{Cross-Modality Decoder Block (SMDB)}

In CMDB, given the two modality feature inputs, first BN is applied to them. Then two cross-modality global feature blocks (CGFB) are used to capture global dependency across the modalities. These processed features are used as residual connections, which are further refined using two CBAMs. Following this step, the features are concatenated in the channel dimension. Then, BN and CB are applied sequentially and used as a residual connection to incorporate local correlation-based feature enhancement.

Figure \ref{fig5} shows the structure of CGFB, which is mainly designed based on our proposed CM-S6 mechanism to capture global dependency across two different modality features. First, two convolution layers are applied to the input features for linear projection, and then the 2D feature maps are rearranged to 1D sequences using the Im2Seq operation. Following this, we apply the CM-S6 mechanism, which is based on a discretized SSM. Algorithm 3 explains the CM-S6 mechanism, which transforms two modality input sequences $f_x\in \mathbb{R}^{L,D}$ and $f_y\in \mathbb{R}^{L,D}$ to the output sequence $f_o\in \mathbb{R}^{L, D}$. The CM-S6 mechanism is similar to S6, the main difference is that the model dynamics is not dependent on the model input $f_y$. Instead, here we derive the learnable parameters $\boldsymbol{B}, \boldsymbol{C}, \boldsymbol{\triangle}$ from the other modality feature $f_x$.  Deriving $\boldsymbol{\triangle}$ from $f_x$ makes the time spent on current input from one modality $f_{y_{k,:}}$ dependent on other modality $f_x$. Moreover, by deriving $\boldsymbol{B}$ and $\boldsymbol{C}$  from $f_x$, our control on how much $f_y$ can be into the $h$, and how much $h$ can be into the output $f_o$ becomes dependent on $f_x$. Thus, the model dynamics become dependent on $f_x$, where the model input is $f_y$. This way, the CM-S6 mechanism can effectively capture the dependency between the two modalities. In addition, similar to SGFB, we also flip the sequence and apply the CM-S6 mechanism for capturing cross-modal global dependency in two directions.

\section{Experiments}
\label{sec4}
\begin{figure*} [h!]
\centering  \includegraphics[width=\linewidth]{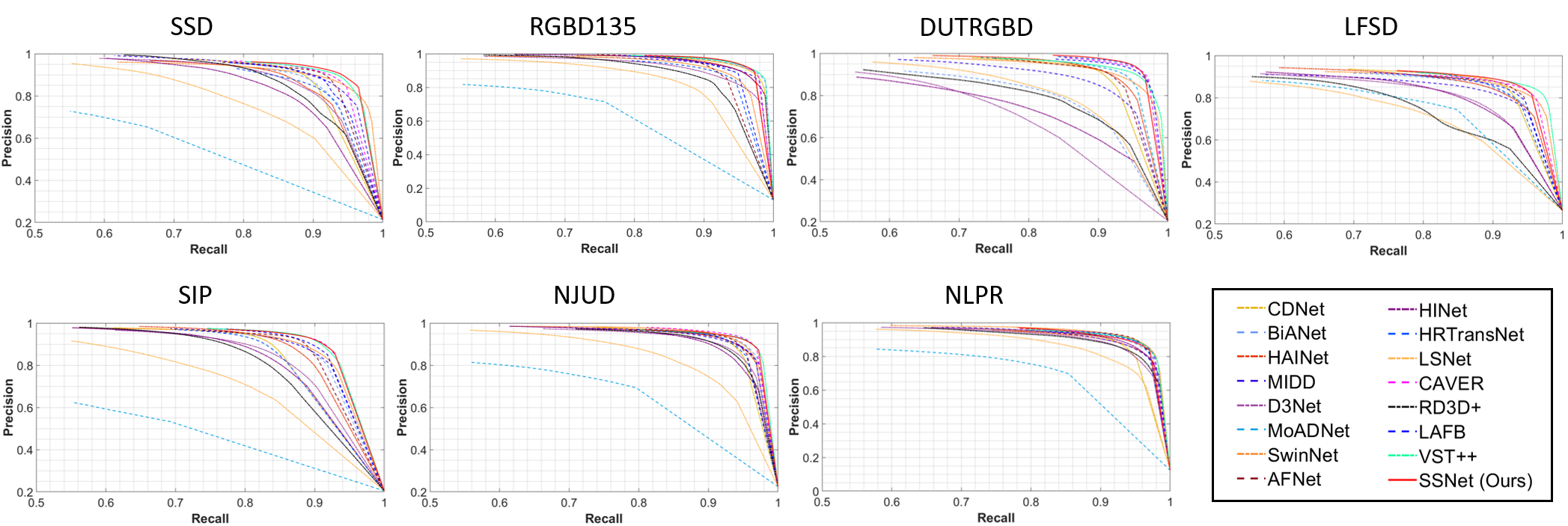}
\caption{PR curves of SSNet with 15 SOTA methods on seven RGB-D SOD datasets.}
\label{tab:sota_pr}
\end{figure*}
\subsection{Setup}
\noindent
\textbf{Datasets:} To evaluate our proposed method, we conducted experiments on seven benchmark datasets: SSD \cite{ssd} ($80$ image pairs), RGBD135 \cite{des} ($135$ image pairs), DUTRGBD \cite{dut} ($1200$ image pairs), LFSD \cite{lfsd} ($100$ image pairs), SIP \cite{sip} ($929$ image pairs), NJUD \cite{sal_trad1} ($1985$ image pairs), and NLPR \cite{nlpr} ($1000$ image pairs). We follow the settings in previous works \cite{caver, disentangled, emtrans, vstplus} and select $1485$ image pairs from NJUD, $700$ from NLPR, and $800$ from DUTRGBD in the training set. The remaining image pairs are used for testing. In summary, our training set contains $2985$ image pairs.

\noindent
\textbf{Training settings:} We train SSNet by minimizing the hybrid loss \cite{loss}. The training is conducted for $100$ epochs with the SGD optimizer with a momentum of $0.9$ and a weight decay of $5\times10^{-4}$. The initial learning rate is $5\times10^{-3}$ and scheduled by cosine strategy. The batch size is set to $8$. The RGB and depth images are resized to $256\times 256$. The single channel depth image is repeated three times along the channel dimension. We employ data augmentation techniques such as color jittering, affine transforms, and horizontal flipping to avoid over-fitting. We conducted all our experiments using NVIDIA A40 GPU within the PyTorch framework.

\noindent
\textbf{Evaluation metrics:} We evaluate our method with four widely used metrics: $MAE$, $S_m$, $F_\beta$, and $E_m$. MAE measures the pixel-level similarity between the saliency map and ground truth (GT). $S_m$ measures the object-aware and region-aware structural similarity. $F_\beta$ considers both precision and recall under an optimal threshold value. $E_m$ measures both pixel-level and image-level errors. Moreover, we plot the precision-recall curves (PR curves) to demonstrate more comprehensive results. We follow \cite{hidanet, spnet} for calculating the evaluation metrics.

\noindent
\textbf{Implementation details:} In SSNet, we use ResNet-101 as backbone networks for extracting features from RGB and depth images. The backbone networks are pretrained with the ImageNet dataset. For the convolution layers in SGFB and CGFB, we set the kernel size to $1\times 1$. For other convolution layers in M\textsuperscript{2}DM and RM, we set the kernel size to $3\times 3$. For S6 and CM-S6 in SGFB and CGFB, we set the state dimension $N$ to $64$. Following \cite{mamba}, we use a parallel algorithm optimized on the CUDA hardware to implement the S6 and CM-S6 mechanisms.

\subsection{Performance Comparison to SOTA Methods}
We compare the performance of SSNet with 15 recent SOTA methods: CDNet \cite{cdnet}, BiANet \cite{bianet}, HAINet \cite{hainet}, MIDD \cite{midd}, D\textsuperscript{3}Net \cite{sip}, MoADNet \cite{moadnet}, SwinNet \cite{swinnet}, AFNet \cite{afnet}, HINet \cite{hinet}, HRTransNet \cite{hrtransnet}, LSNet \cite{lsnet}, CAVER \cite{caver}, RD3D+ \cite{rd3dplus}, LAFB \cite{lafb}, and VST++ \cite{vstplus}. For the SOTA methods, we use their trained network parameters and official code to generate the saliency maps.

\begin{figure} [h!]
\centering  \includegraphics[width=\linewidth]{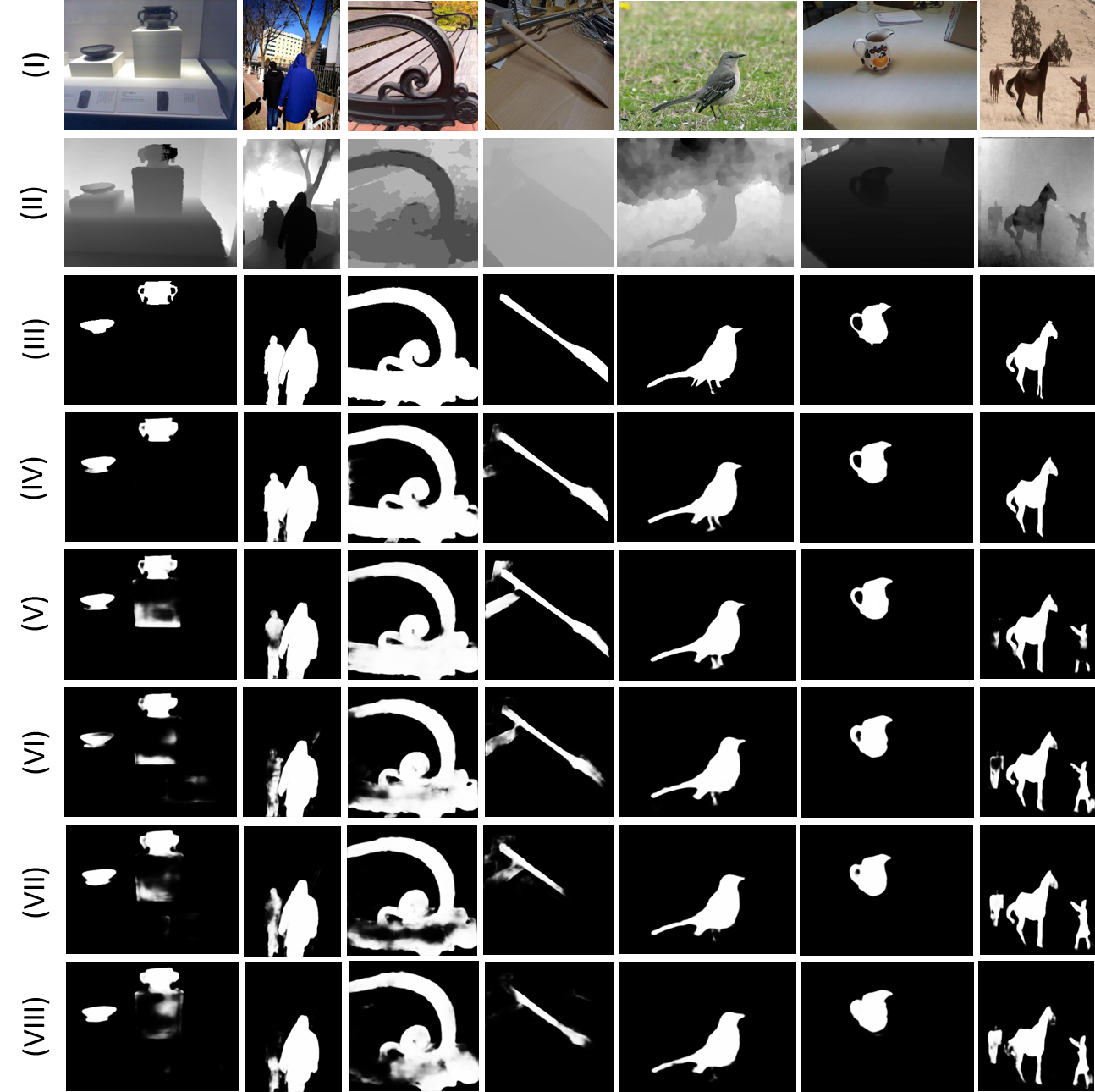}
\caption{Visual comparisons to ablation experiments for different components. (I) RGB image, (II) depth images, (III) GT, (IV) SSNet (Ours), (V) Baseline+ACE+SEM+SMDB, (VI) Baseline+ACE+SEM, (VII) Baseline+ACE, (VIII) Baseline.}
\label{fig:ablations_visual}
\end{figure}
\noindent
\textbf{Quantitative Comparison:} Table \ref{tab:sota} presents the quantitative comparison of RGB-D SOD task on seven benchmark datasets. Along with the four metrics $MAE$, $S_m$, $F_\beta$, and $E_m$, we also tabulate the no. of model parameters, FLOPs count, and average runtime on a GPU. We report the values of FLOPs and runtime under the setting of SOD from RGB-D images with a resolution of $384\times 384$. Except for the NLPR dataset, our method achieves leading performance across the other six datasets. Among the SOTA methods, CAVER and VST++ show performance comparable to our method. SSNet has a lower parameter count and FLOPs count compared to CAVER and VST++. Moreover, we plot the PR curves for the seven datasets in Figure \ref{tab:sota_pr} for a more comprehensive evaluation. The curves corresponding to our method are positioned more upward, indicating superior performance.

\noindent
\textbf{Qualitative Comparison:} Figure \ref{fig:sota_visual} presents visual comparisons of SSNet with several recent SOTA methods across different complex scenes. As demonstrated by the results, compared to SOTA methods, our method more effectively detects salient objects even when the foreground and background share similar appearances. This can be attributed to our SSM-based decoder design, which enhances scene understanding by modeling global context across different modalities, along with our saliency-aware SEM. Additionally, our ACE module enhances low-quality depth maps, making them a more reliable source of information.

\subsection{Ablation Study}

The performance of our proposed method depends on the network design of SSNet and the ACE stage. In particular, our proposed SSM-based SMDB and CMDB effectively capture the global dependency in intra and inter-modalities. Moreover, the ACE stage enhances the contrast of low-quality depth maps to make them more useful for the SOD task. Additionally, our proposed SEM integrates saliency priors with the deep features from RGB-D images, improving their discriminative capability. In this section, we conduct an ablation study to evaluate the effectiveness of our proposed components. Table \ref{tab:ablation} presents the quantitative results on seven datasets and Figure \ref{fig:ablations_visual} shows the visual comparison of ablation experiments. We first create a baseline model, denoted as "Baseline", by removing ACE, SEM, SMDB, and CMDB from our proposed method. 

\noindent
\textbf{Effectiveness of CMDB:} Our proposed SSM-based CMDB is designed to capture global dependency between different modality feature maps. To show its effectiveness, we remove it from SSNet and conduct experiments. Comparison between SSNet ("Baseline+ACE+SEM+SMDB+CMDB") and "Baseline+ACE+SEM+SMDB" shows that removing CMDB significantly degrades the SOD performance in most of the cases.

\noindent
\textbf{Effectiveness of SMDB:} Our proposed SSM-based SMDB is designed to capture global dependency in self-modality feature maps. Comparison between "Baseline+ACE+SEM" and "Baseline+ACE+SEM+SMDB" shows that the results improve when SMDB is used.

\noindent
\textbf{Effectiveness of SEM:} Our novel SEM integrates three saliency priors with deep features to improve their discriminative capability. As evident from the comparison between "Baseline+ACE" and "Baseline+ACE+SEM", removing SEM degrades the SOD performance.

\noindent
\textbf{Effectiveness of ACE:} Our proposed ACE scheme improves the contrast of low-quality depth maps.
To show the effectiveness of ACE, we remove it and feed the depth maps directly for saliency prior calculation and to SSNet. Comparison of results between "Baseline" and "Baseline+ACE" shows that using ACE significantly improves the SOD performance.

The quantitative and visual comparison results of the ablation experiments show that all components play an important role in SOD performance, which degrades to varying degrees when any component is removed. These findings demonstrate the overall effectiveness of our proposed method.
\begin{table} [h!]
\centering
\caption{Ablation experiments of different components. Best values are highlighted. $\uparrow$ and $\downarrow$ mean high and low values are desired, respectively.}
  \includegraphics[width=\linewidth]{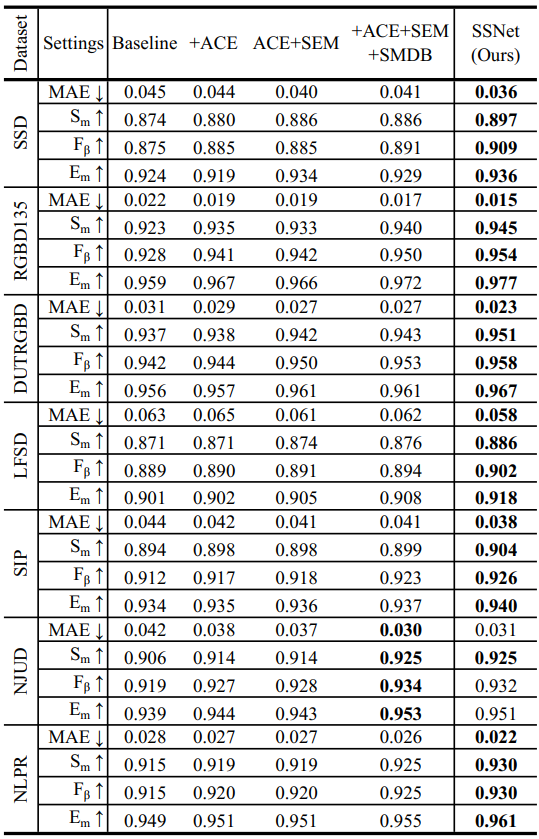}
\label{tab:ablation}
\end{table}
\begin{figure} [h!]
\centering  \includegraphics[width=\linewidth]{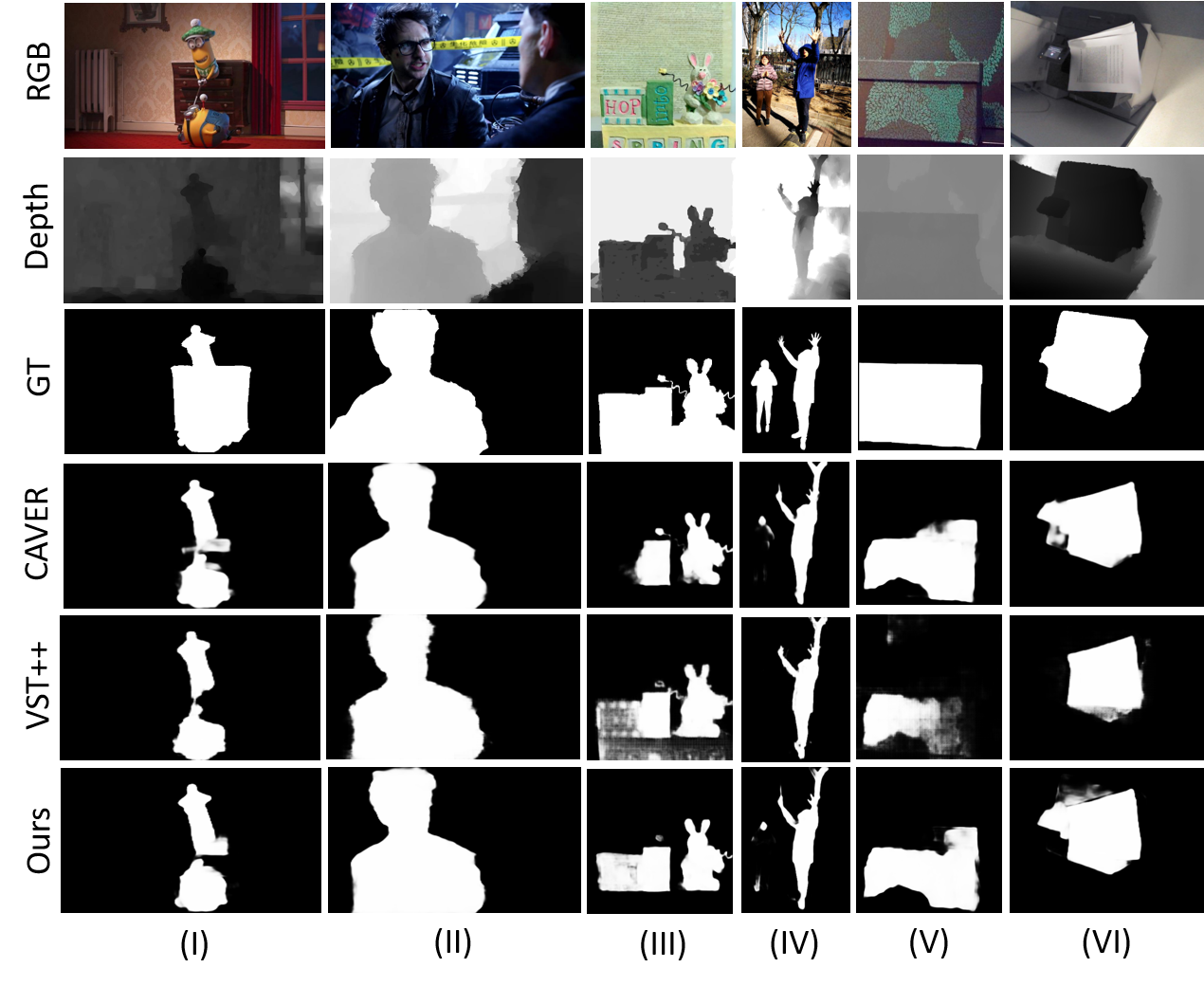}
\caption{Typical failure cases of SSNet and two recently proposed SOTA methods.}
\label{fig:failures}
\end{figure}
\subsection{Failure Cases}
Figure \ref{fig:failures} illustrates several typical failure cases of SSNet, alongside two recently proposed SOTA methods. In some instances, SSNet struggles with object boundaries and background differentiation. For example, in Case (I), certain parts of a body resting on a chair are misclassified as background. In Case (II), while SSNet correctly detects the body silhouette in alignment with the depth map, the ground truth (GT) image appears misleading. Case (III) demonstrates another challenge, where SSNet successfully detects objects on a table but fails to recognize the table itself as a salient object. In more complex scenarios, SSNet faces additional limitations. Case (IV) highlights an issue with noisy depth maps, where only one person and the tree stem are captured. As a result, SSNet fails to detect the second person present in the RGB image and mistakenly classifies the tree stem as salient. Case (V) presents another difficulty, where the GT image defines an object as salient despite it not meeting standard visual saliency criteria, leading to incorrect detection. Finally, Case (VI) involves an extremely complex scene, making it difficult to segment the salient object, causing SSNet to fail. Despite these challenges, it is important to note that other SOTA methods also struggle in these cases, highlighting the inherent complexity of RGB-D SOD tasks.
\section{Conclusion}
\label{sec5}
In this work, we introduce an SSM and saliency prior-based network for the RGB-D SOD task. Specifically, we propose a novel SSM-based CM-S6 mechanism to capture global dependency between different modalities and an SEM to integrate saliency priors with deep features to improve their discriminative capability. Moreover, our proposed ACE enhances the contrast of low-quality depth maps to make them more suitable for the SOD task. We conducted experiments across seven benchmark datasets, demonstrating that our method outperforms existing SOTA approaches, as evidenced by both quantitative metrics and qualitative results. Furthermore, our ablation study validates the effectiveness of each proposed component in improving saliency detection performance. Future work could explore the effects of our proposed SSM-based CM-S6 mechanism on different downstream tasks in cross-modality settings.

\bibliographystyle{IEEEtran}
\bibliography{main}
\end{document}